\pdfoutput=1
\documentclass[11pt]{article}
\usepackage[table]{xcolor}

\usepackage[preprint]{acl}

\usepackage{times}
\usepackage{latexsym}

\usepackage[T1]{fontenc}
\usepackage[utf8]{inputenc}

\usepackage{microtype}

\usepackage{inconsolata}

\usepackage{graphicx}

\usepackage{fancyvrb, fvextra}
\usepackage{dsfont}
\usepackage{amsmath}
\usepackage{multirow}

\title{Does Using Counterfactual Help LLMs Explain \\ Textual Importance in Classification?}

\author{
 \textbf{Nelvin Tan\textsuperscript{1}},
 \textbf{James Asikin Cheung\textsuperscript{1}},
 \textbf{Yu-Ching Shih\textsuperscript{1}},
 \textbf{Dong Yang\textsuperscript{1}},
 \textbf{Amol Salunkhe\textsuperscript{1}},
\\
\\
 \textsuperscript{1}American Express
 \\
 \tiny{
   \textbf{Correspondence:} \href{thongcainelvin.tan@aexp.com}{thongcainelvin.tan@aexp.com}
 }
}

\begin{document}

\maketitle

\begin{abstract}
Large language models (LLMs) are becoming useful in many domains due to their impressive abilities that arise from large training datasets and large model sizes. More recently, they have been shown to be very effective in textual classification tasks, motivating the need to explain the LLMs' decisions. Motivated by practical constrains where LLMs are black-boxed and LLM calls are expensive, we study how incorporating counterfactuals into LLM reasoning can affect the LLM's ability to identify the top words that have contributed to its classification decision. To this end, we introduce a framework called the \textit{decision changing rate} that helps us quantify the importance of the top words in classification. Our experimental results show that using counterfactuals can be helpful.
\end{abstract}

\section{Introduction}

Large language models (LLMs) are becoming useful in many domains \cite{Li2023a,Sakai2025,Tan2025} due to their impressive abilities that arise from large training datasets \cite{Li2023b} and large model sizes \cite{Naveed2023}. More recently, they have been shown to be very effective in textual classification tasks \cite{Zhao2024a,Sakai2025}, motivating the need to explain the LLMs' decisions. In this paper, we investigate how well LLMs can explain their decisions when making textual classifications -- specifically, we study the LLM's ability to select the top $k$ key words that led to its classification decision. Some of the potential business use cases of being able to identify the keys words that lead to a certain LLM classification are as follows: (i) Can aid the prompt designer to trim unneccesary words or shorten phrases. This reduces the number of tokens processed which helps to reduce cost. (ii) Helps to provide auditable records of how text inputs drive LLM classifications.

\paragraph{Conterfactuals.} In this paper, we focus on a specific type of self-generated explanations called self-generated counterfactual explanations \cite{Wachter2017,Dehghanighobadi2025} -- which we call simply as counterfactuals: Given an input $\boldsymbol{x}$ and the LLM output $\hat{y}$, the counterfactual $\boldsymbol{x}'$ is a minimally modified version (i.e., the edit/Levenshtein distance between $\boldsymbol{x}$ and $\boldsymbol{x}'$ are minimized) of the input such that when $\boldsymbol{x}'$ is feed into the LLM, its output $\hat{y}\neq\hat{y}'$. For example, $\boldsymbol{x}$ is \textit{`This item is neither affordable, nor of good quality.'} and $\boldsymbol{x}'$ is \textit{`This item is both affordable, and of good quality.'} 

Note that we do not solve any optimization problem to obtain $\boldsymbol{x}'$, but simply ask for the LLM for $\boldsymbol{x}'$. This is motivated by recent work \cite{Dehghanighobadi2025} that showed that LLMs are generally able to produce counterfactuals with a high success rate for classification problems with a few classes. We build on their work to study how these counterfactuals affect key-word identificiation.

\paragraph{Problem setup.} Given an input $\boldsymbol{x}$ and  completely black-box LLM (i.e., no accesss to weights, gradients, etc.), we want to produce the top-$k$ words that explain the classification decision of the LLM, where $k$ is up to the user to choose.

\paragraph{Motivation.} Our problem setup is motivated by businesses that use black-box LLMs (e.g., GPT-4o) but have no access to the inner workings of the LLMs. At the same time, LLM calls are expensive so we want to minimize it for cost savings. This leads to incoporating counterfactuals into LLM reasoning as a viable approach for LLM explainability. Our main research questions are:
\begin{enumerate}
    \item Do counterfactuals help LLM with explaining the importance of words in textual classification?
    \item How do we score the LLM's ability to select the top-$k$ words that influenced its classification decision?
\end{enumerate}

\section{Related Work}
Known black-box explainability techniques for textual classification and for LLMs are as follows:
\begin{itemize}
	\item SHAP \cite{Lundberg2017} and LIME \cite{Ribeiro2016} are common input-feature based techniques that works reasonably well for smaller models, but are they too expensive for LLMs, requiring 100-1000 LLM calls per classification output which is not practical due to the high cost. Furthermore, prior work \cite{Ichikawa2014} argues that counterfactuals constitute a better test of knowledge over LIME and SHAP.
	\item Prompting the LLM directly can take as low as 1 prompt, but this totally ignores the internal mechanism of the LLM and it has been shown that LLMs do not always say what they think \cite{Chen2025}. Nevertheless, it has been shown to align with human intuition \cite{Agarwal2024}. There has also been other work that has argued that the contrastive nature of counterfactuals are better aligned with human reasoning \cite{Miller2019} -- making it interesting to study whether counterfactuals can improve on direct prompting.
\end{itemize}
Black-box approaches aside, there are many other LLM explainability approaches -- we refer the interested reader to the following surveys \cite{Zhao2024b,Luo2024} for a comprehensive review.

\paragraph{Main contributions.} With respect to the main research questions, are contributions are as follows:
\begin{enumerate}
    \item We investigate the usefuless of counterfactuals in helping LLM explain the importance of input words in textual classification. Our experimental results show that using counterfactuals can help improve the identification of the top-$k$ words.
    \item To achieve the above goal, we introduce a metric known as the \textit{decision-changing} rate, which captures how important the top-$k$ words to the LLM's classification decision. To the best of our knowledge, this is a new metric that could be useful for the field.
\end{enumerate}

\section{Methodology}

We introduce 3 approaches to obtain the top-$k$ words that led to an LLM's classification decision: \textit{direct prompting}, \textit{counterfactual-parallel}, and \textit{counterfactual-sequential}. The prompts for all 3 approaches are given in Appendices \ref{app:amazon_prompts}, \ref{app:SST2_prompts}, and \ref{app:IMDB_prompts}, and we give the high-level ideas below.

\begin{figure}[t]
    \centering
    \includegraphics[width=1.0\columnwidth]{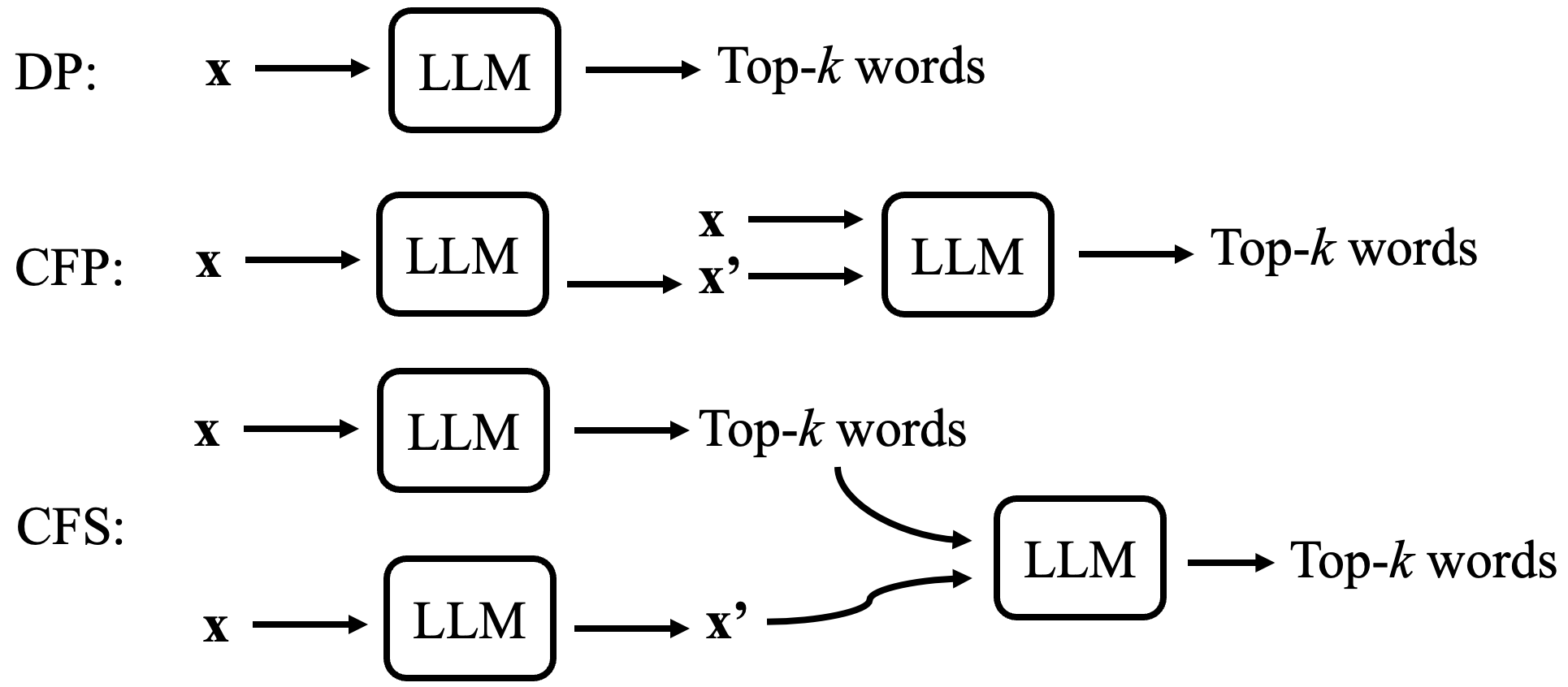}
    \caption{Visualization of the different approaches}
    \label{fig:approaches}
\end{figure}

\paragraph{Direct prompting (DP).} Directly prompt the LLM for the classification and ask it to list the top-$k$ most important words in the input that led to its classification decision. This is shown visually in Figure \ref{fig:approaches}.

\paragraph{Counterfactual-parallel (CFP).} The key steps are shown visually in Figure \ref{fig:approaches} and are stated as follows: (1.) Ask the LLM to produce the counterfactual $\boldsymbol{x}'$. (2.) Check if the LLM outputs satisfy $\hat{y}'\neq\hat{y}$; if this is not true then use DP to produce the top-$k$ most important words. Otherwise, proceed with the next step. (3. Parallel step) Feed $\boldsymbol{x}$ and $\boldsymbol{x}'$ into the LLM and ask it to produce the top-$k$ most important words that led to its previous classification decision $\hat{y}$.

\paragraph{Counterfactual-sequential (CFS).} The key steps are shown visually in Figure \ref{fig:approaches} and are stated as follows: (1.) Use DP to get the top-$k$ most important words that led to its classification decision $\hat{y}$. (2.) Ask the LLM to produce the counterfactual $\boldsymbol{x}'$. (3.) Check if the LLM outputs satisfy $\hat{y}'\neq\hat{y}$; if this is not true then use DP to produce the top-$k$ most important words. Otherwise, proceed with the next step. (4. Refinement step) Feed $\boldsymbol{x}$ and $\boldsymbol{x}'$ into the LLM and ask it to refine its top-$k$ most important words that led to its classification decision. It is allowed to keep the initial top $k$ words as the final choice of words if it is confident.

\paragraph{Extensions to CFP and CFS.} We can take advantage of multiple LLM prompts (i.e., sampling) to produce a weights vector corresponding to the words in the input text, where each word in the input text corresponds to a weight and the weight indicates how important the word is in the classification decision. The key steps are as follows: (i) Set the temperature of the LLM to be greater than 0 (e.g., 1), and decide on the number of important words $k$. (ii) Have a vector of weights $\boldsymbol{w}$ (initialized to all zeros) with each entry corresponding to a word in $\boldsymbol{x}$. (iii) Run either CFP or CFS $n>1$ times. For each run, if a word is included in the top-$k$ most important words, then increment its corresponding weight in $\boldsymbol{w}$ by 1. (iv) Divide all entries in $\boldsymbol{w}$ by $n$ and produce $\boldsymbol{w}$ as the final output. This vector $\boldsymbol{w}$ can be used to produce a heatmap of the words.

\section{Experiments} \label{sec:experiments}

\subsection{Dataset and Evaluation Metric}

We look at the following textual binary classification datasets of different lengths:
\begin{itemize}
    \item \textbf{Amazon \cite{Kotzias2015} (short-length).} Dataset consists of 500 positive and 500 negative sentences from customer reviews were extracted from the \url{amazon.com}. 
    \item \textbf{SST2 \cite{Pang2005} (medium-length).} Dataset consists of 11,855 single sentences extracted from movie reviews.
    \item \textbf{IMDB \cite{Maas2011} (long-length).} Dataset consists of 50,000 reviews from the Internet Movie Database (IMDb) labeled as positive or negative.
\end{itemize}
We use the following LLMs with temperature set to 0 for consistency of results: LLaMA3-70B (weaker model) and GPT-4o (stronger model). For each dataset, we randomly sample 100 reviews for our experiments -- for our IMDB dataset, whenever the review exceeds the context length of LLaMA3-70B, we manually modify it to fit into the context length (we also run GPT-4o on this modified dataset). Details of our sampled dataset are as follows:
\begin{itemize}
	\item \textbf{Amazon.} Average of 10.86 words per review; our aim would be to seek the top-$k$ most important words, where $k=1,2,3$ (9.21\%, 18.42\%, 27.62\% correspondingly).
	\item \textbf{SST2.} Average of 17.76 words per review; our aim would be to seek the top-$k$ most important words, where $k=1,2,3$ (5.63\%, 11.26\%, 16.89\% correspondingly).
	\item \textbf{IMDB.} Average of 213.28 words per review; our aim would be to seek the top-$k$ most important words, where $k=3,5$ (1.41\%, 2.34\% correspondingly).
\end{itemize}
We run the following approaches: CFP, CFS, and DP. For each approach, the same LLM is used throughout the entire approach. The prompts for the experiments are given in Appendices \ref{app:amazon_prompts} (Amazon dataset), \ref{app:SST2_prompts} (SST2 dataset), and \ref{app:IMDB_prompts} (IMDB dataset).

\paragraph{Decision changing rate.} We introduce a metric to measure explainability: \textit{decision-changing} rate (DCR), which is the average of the decision-changing scores defined below:
\begin{itemize}
	\item For a given list of important words $\mathcal{S}$ and input text $\boldsymbol{x}$, for each word $w\in\mathcal{S}$, replace it in $\boldsymbol{x}$ with `\{MASK\}'.
	\item Ask the LLM (that initially gave the classification), to produce a modified review $\boldsymbol{x}''$ by replacing each sub-string `\{MASK\}' with a word or short phrase such that the overall sentiment of the movie review gets flipped. The LLM is also asked to make sure that no other replacements are made, and that the review flows coherently.
	\item Feed $\boldsymbol{x}''$ back into the LLM and ask for its classification. If the classification is flipped, then we set the decision-changing score to be 1. Otherwise, we set it to be 0.
\end{itemize}

\subsection{Main Results} \label{sec:main_results}

\begin{table*}
    \centering
    \begin{tabular}{ccccccccccccc}
      \hline
      \multirow{2}{*}{Approach} & \multicolumn{4}{c}{\textbf{Amazon (short)}} & \multicolumn{4}{c}{\textbf{SST2 (medium)}} & \multicolumn{2}{c}{\textbf{IMDB (long)}} \\
       & Acc. & $\text{DCR}_1$ & $\text{DCR}_2$ & $\text{DCR}_3$ & Acc. & $\text{DCR}_1$ & $\text{DCR}_2$ & $\text{DCR}_3$ & Acc. & $\text{DCR}_3$ & $\text{DCR}_5$ \\
      \hline
      CFP-L3 & 98\% & \cellcolor{blue!10}0.82 & \cellcolor{blue!10}0.92 & \cellcolor{blue!10}0.96 & 94\% & \cellcolor{blue!10}0.71 & \cellcolor{blue!10}0.90 & 0.91 & 99\% & 0.70 & 0.75 \\
      CFS-L3 & 98\% & 0.74 & 0.88 & 0.93 & 96\% & \cellcolor{blue!10}0.71 & 0.84 & \cellcolor{blue!10}0.92 & 99\% & \cellcolor{blue!10}0.71 & \cellcolor{blue!10}0.80 \\
      DP-L3 & 98\% & \cellcolor{blue!10}0.82 & 0.90 & \cellcolor{blue!10}0.96 & 96\% & 0.70 & 0.82 & 0.91 & 99\% & 0.65 & 0.72 \\
      CFP-G4 & 98\% & \cellcolor{blue!10}0.69 & \cellcolor{blue!10}0.88 & \cellcolor{blue!10}0.93 & 99\% & \cellcolor{blue!10}0.69 & 0.82 & 0.83 & 98\% & \cellcolor{blue!10}0.46 & \cellcolor{blue!10}0.70 \\
      CFS-G4 & 98\% & 0.68 & 0.87 & \cellcolor{blue!10}0.93 & 99\% & 0.65 & \cellcolor{blue!10}0.84 & \cellcolor{blue!10}0.85 & 96\% & \cellcolor{blue!10}0.46 & 0.69 \\
      DP-G4 & 98\% & \cellcolor{blue!10}0.69 & 0.87 & \cellcolor{blue!10}0.93 & 99\% & 0.62 & 0.77 & 0.81 & 96\% & 0.44 & 0.68 \\
      \hline
    \end{tabular}
    \caption{L3 refers to LLaMA3-70B and G4 refers to GPT-4o. The best DCR for each LLM and dataset is highlighted in blue. The subcript of DCR denotes $k$. Acc.~refers to the average accuracy over the $k$'s.}
    \label{tab:main_result}
\end{table*}

\begin{figure}[t]
    \centering
    \includegraphics[width=1.0\columnwidth]{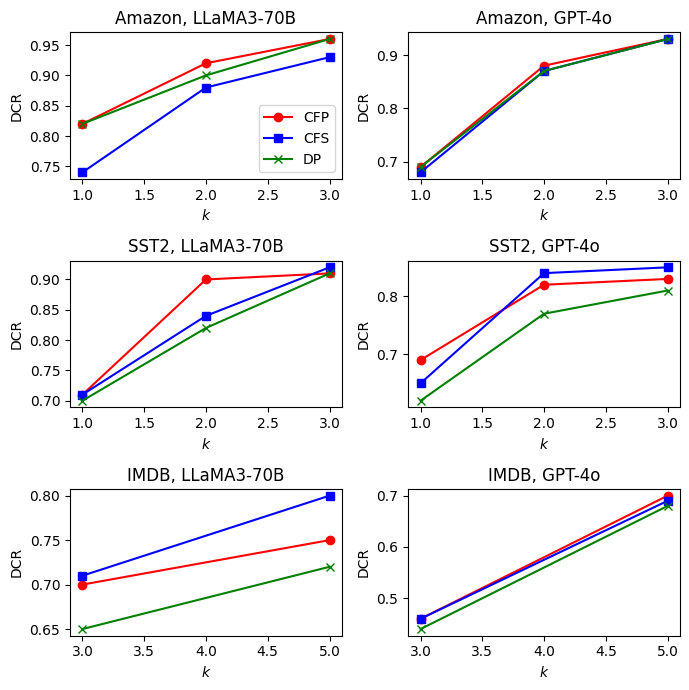}
    \caption{Visualization of the different approaches}
    \label{fig:k_plots}
\end{figure}

Our experimental results are displayed in Table \ref{tab:main_result} and Figure \ref{fig:k_plots}. As expected, the accuracy generally remains unchanged for a given dataset and LLM, and increasing $k$ increases the DCR since there are more opportunities for the LLM to change the sentiment of the review. Furthermore, the DCR for LLaMA3-70B (weaker model) is generally higher than that of GPT-4o (stronger model) because a weaker model's classification decision is more easily swayed due to its weaker reasoning abilities. 

We also observe that the smaller the proportion of $k$ (over the word length of the input string), the lower the DCR -- we can observe this by looking at how $\text{DCR}_3$ decreases across Amazon (short), SST2 (medium), and IMDB (long). This is because when proportion of $k$ is bigger (like in Amazon), there are few words/phrases outside of the top-$k$ words that can counter the effect of the replaced words. However, when the proportion of $k$ is smaller (like in IMDB), there are many words/phrases outside of the top-$k$ words that can counter the effect of the replaced words. For example, consider a movie review with $k=3$ and a negative sentiment, the review \textit{`This movie is \{MASK\}, \{MASK\}, and \{MASK\}.'} is much easier to negate (to become positive) compared to the longer review \textit{`This movie is \{MASK\}, \{MASK\}, and \{MASK\} -- I am definitely not recommending it to anyone!'}

Finally, we have the following observations from Table \ref{tab:main_result} and Figure \ref{fig:k_plots} for each dataset:
\begin{itemize}
    \item For short-length inputs (Amazon), we observe that CFP performs the best, outperforming or matching DP's performances, whereas CFS performs equal to or worse than DP. 
    \item For medium-length inputs (SST2), both CFP and CFS outperform DP for all cases, with CFP being the best performer $3$ times and CFS being the best performer $4$ times. 
    \item For long-length inputs (IMDB), both CFP and CFS outperform DP for all cases, with CFP being the best performer $2$ times and CFS being the best performer $3$ times. 
\end{itemize}
Overall, CFP is able to consistently outperform or match DP's performance. While CFS is also able to outperform and match DP's performance in most cases, there were a few cases where it underperformed, making it slightly inconsistent. Therefore, we would recommended the use of CFP over CFS or DP.

\section{Conclusion}

Our study shows that incorporating counterfactuals into LLM key-words identification can be useful under the framework that we have introduced. We can further extend our approach to multi-class classification problems by reducing them into two classes -- e.g., (positive, neutral, negative) can be reduced to (positive, not positive). Specifically, if the predicted sendiment is positive, then we run CFP on the classes (positive, not positive) and the output would be the top-$k$ words that led to a positive classification. We leave the exploration of this idea as a future work.

\section{Limitations}

While we believe that our choice of LLM should be sufficient, we acknowledge that we did not exhaustively evaluate a large selections of LLMs. Regarding the decision-changing score, we notice that despite our best efforts to be explicit to the LLM (and also setting the temperature to 0) when asking it to \textit{only} replace the masked words, the LLM sometimes (but rarely) replace some other non-mask words. This injects some minor noise to the DCR calculation.

% \section{Acknowledgments}

% While this work was conducted at American Express, the conclusions and views expressed are those of the authors and do not necessarily reflect the views of the company.

\newpage

\bibliography{custom}

\appendix

\newpage

% \onecolumn

\section{Prompt for Amazon Dataset} \label{app:amazon_prompts}

We provide the prompts used for the Amazon dataset. The user's prompts are colored in red, and the LLM's outputs are colored in black. We present the case of $k=3$.

\paragraph{Prompts for DP:}

\begin{Verbatim}[formatcom=\color{red},breaklines,fontsize=\tiny,frame=single]
Given the customer review '{review}', your task is to classify it either as 'positive' or 'negative', and state the top 3 most important words that led to you decision. Please follow the following output format and produce nothing else: <class,word1,word2,word3>
\end{Verbatim}

\begin{Verbatim}[breaklines,fontsize=\tiny,frame=single]
<classification1,DP word1,DP word2,DP word3>
\end{Verbatim}

\paragraph{Prompts for CFP:}

\begin{Verbatim}[formatcom=\color{red},breaklines,fontsize=\tiny,frame=single]
Given the customer review '{review}', your task is to classify it either as 'positive' or 'negative' (and nothing else). Please only produce the classification and enclose it within <new> and </new> tags (i.e., <new>positive</new> or <new>negative</new>).
\end{Verbatim}

\begin{Verbatim}[breaklines,fontsize=\tiny,frame=single]
<new>{classification1}</new>
\end{Verbatim}

\begin{Verbatim}[formatcom=\color{red},breaklines,fontsize=\tiny,frame=single]
Customer reviews are classified as either positive or negative. The customer review '{review}' is classified as 'negative'. Your task is to modify the review with minimal edits to flip the sentiment prediction. Please only produce the modified review and enclose it within <new> and </new> tags. Avoid making any unnecessary changes, and don't produce anything else.
\end{Verbatim}

\begin{Verbatim}[breaklines,fontsize=\tiny,frame=single]
<new>{counterfactual}</new>
\end{Verbatim}

\begin{Verbatim}[formatcom=\color{red},breaklines,fontsize=\tiny,frame=single]
Given the customer review '{counterfactual}', your task is to classify it either as 'positive' or 'negative' (and nothing else). Please only produce the classification and enclose it within <new> and </new> tags.
\end{Verbatim}

\begin{Verbatim}[breaklines,fontsize=\tiny,frame=single]
<new>{classification2}</new>
\end{Verbatim}

\begin{Verbatim}[formatcom=\color{red},breaklines,fontsize=\tiny,frame=single]
A counterfactual of a customer review x1, is another customer review x2 that is modified from x1 with a minimal number of edits such that the classification of x2 is different from x1. Given a customer review x1 that states <{review}> and its counterfactual x2 that states <{counterfactual}>, your task is to study x1 and x2 carefully, and state the top 3 most important words (that are not stop words) from x1 that led to your previous decision of '{classification1}'. Please follow the following output format and produce nothing else: [word1,word2,word3]
\end{Verbatim}

\begin{Verbatim}[breaklines,fontsize=\tiny,frame=single]
{top-3 words}
\end{Verbatim}

\paragraph{Prompts for CFS:}

\begin{Verbatim}[formatcom=\color{red},breaklines,fontsize=\tiny,frame=single]
Given the customer review '{review}', your task is to classify it either as 'positive' or 'negative', and state the top 3 most important words that led to you decision. Please follow the following output format and produce nothing else: <class,word1,word2,word3>
\end{Verbatim}

\begin{Verbatim}[breaklines,fontsize=\tiny,frame=single]
<classification1,DP word1,DP word2,DP word3>
\end{Verbatim}

\begin{Verbatim}[formatcom=\color{red},breaklines,fontsize=\tiny,frame=single]
Customer reviews are classified as either positive or negative. The customer review '{review}' is classified as '{classification1}'. Your task is to modify the review with minimal edits to flip the sentiment prediction. Please only produce the modified review and enclose it within <new> and </new> tags. Avoid making any unnecessary changes, and don't produce anything else.
\end{Verbatim}

\begin{Verbatim}[breaklines,fontsize=\tiny,frame=single]
<new>{counterfactual}</new>
\end{Verbatim}

\begin{Verbatim}[formatcom=\color{red},breaklines,fontsize=\tiny,frame=single]
Given the customer review '{counterfactual}', your task is to classify it either as 'positive' or 'negative' (and nothing else). Please only produce the classification and enclose it within <new> and </new> tags.
\end{Verbatim}

\begin{Verbatim}[breaklines,fontsize=\tiny,frame=single]
<new>{classification2}</new>
\end{Verbatim}

\begin{Verbatim}[formatcom=\color{red},breaklines,fontsize=\tiny,frame=single]
A counterfactual of a customer review x1, is another customer review x2 that is modified from x1 with minimal number of edits such that the classification of x2 is different from x1. Review your previous selection of the 3 most important words in x1 below with relation to its classification (also stated below), using the initial customer review x1 and the counterfactual customer review x2. If you are confident about your answer, maintain your answer. Otherwise, update your answer. Note that the words must only come from x1, and never from x2. Please follow the following output format and produce nothing else: [word1,word2,word3]
### initial customer review x1 with rating '{classification1}'
{review}
### counterfactual customer review x2 with rating '{classification2}'
{counterfactual}
### 3 most important words from x1
[DP word1,DP word2,DP word3]
\end{Verbatim}

\begin{Verbatim}[breaklines,fontsize=\tiny,frame=single]
[CF word1,CF word2,CF word3]
\end{Verbatim}

\paragraph{Prompts for obtaining decision-changing score:}

\begin{Verbatim}[formatcom=\color{red},breaklines,fontsize=\tiny,frame=single]
Given the review '{masked review}', for each sub-string '{MASK}' in the review, replace it with a word or short phrase such that the overall sentiment of the review becomes 'positive'. Also make sure that the sentence flows coherently. Please only produce the modified review and enclose it within <new> and </new> tags. Avoid making any unnecessary changes, never change any part of the review that is not '{MASK}', and don't produce anything else.
\end{Verbatim}

\begin{Verbatim}[breaklines,fontsize=\tiny,frame=single]
<new>{new review}</new>
\end{Verbatim}

\begin{Verbatim}[formatcom=\color{red},breaklines,fontsize=\tiny,frame=single]
Given the customer review '{new review}', your task is to classify it either as 'positive' or 'negative' (and nothing else). Please only produce the classification and enclose it within <new> and </new> tags (i.e., <new>positive</new> or <new>negative</new>).
\end{Verbatim}

\begin{Verbatim}[breaklines,fontsize=\tiny,frame=single]
<new>{new classification}</new>
\end{Verbatim}

\section{Prompt for SST2 Dataset} \label{app:SST2_prompts}

We provide the prompts used for the SST2 dataset. The user's prompts are colored in red, and the LLM's outputs are colored in black. We present the case of $k=3$.

\paragraph{Prompts for DP:}

\begin{Verbatim}[formatcom=\color{red},breaklines,fontsize=\tiny,frame=single]
Given the movie review '{review}', your task is to classify it either as 'positive' or 'negative', and state the top 3 most important words that led to you decision. Please follow the following output format and produce nothing else: <class,word1,word2,word3>
\end{Verbatim}

\begin{Verbatim}[breaklines,fontsize=\tiny,frame=single]
<classification1,DP word1,DP word2,DP word3>
\end{Verbatim}

\paragraph{Prompts for CFP:}

\begin{Verbatim}[formatcom=\color{red},breaklines,fontsize=\tiny,frame=single]
Given the movie review '{review}', your task is to classify it either as 'positive' or 'negative' (and nothing else). Please only produce the classification and enclose it within <new> and </new> tags (i.e., <new>positive</new> or <new>negative</new>).
\end{Verbatim}

\begin{Verbatim}[breaklines,fontsize=\tiny,frame=single]
<new>{classification1}</new>
\end{Verbatim}

\begin{Verbatim}[formatcom=\color{red},breaklines,fontsize=\tiny,frame=single]
Movie reviews are classified as either positive or negative. The movie review '{review}' is classified as 'negative'. Your task is to modify the review with minimal edits to flip the sentiment prediction. Please only produce the modified review and enclose it within <new> and </new> tags. Avoid making any unnecessary changes, and don't produce anything else.
\end{Verbatim}

\begin{Verbatim}[breaklines,fontsize=\tiny,frame=single]
<new>{counterfactual}</new>
\end{Verbatim}

\begin{Verbatim}[formatcom=\color{red},breaklines,fontsize=\tiny,frame=single]
Given the movie review '{counterfactual}', your task is to classify it either as 'positive' or 'negative' (and nothing else). Please only produce the classification and enclose it within <new> and </new> tags.
\end{Verbatim}

\begin{Verbatim}[breaklines,fontsize=\tiny,frame=single]
<new>{classification2}</new>
\end{Verbatim}

\begin{Verbatim}[formatcom=\color{red},breaklines,fontsize=\tiny,frame=single]
A counterfactual of a movie review x1, is another movie review x2 that is modified from x1 with a minimal number of edits such that the classification of x2 is different from x1. Given a movie review x1 that states <{review}> and its counterfactual x2 that states <{counterfactual}>, your task is to study x1 and x2 carefully, and state the top 3 most important words (that are not stop words) from x1 that led to your previous decision of '{classification1}'. Please follow the following output format and produce nothing else: [word1,word2,word3]
\end{Verbatim}

\begin{Verbatim}[breaklines,fontsize=\tiny,frame=single]
{top-3 words}
\end{Verbatim}

\paragraph{Prompts for CFS:}

\begin{Verbatim}[formatcom=\color{red},breaklines,fontsize=\tiny,frame=single]
Given the movie review '{review}', your task is to classify it either as 'positive' or 'negative', and state the top 3 most important words that led to you decision. Please follow the following output format and produce nothing else: <class,word1,word2,word3>
\end{Verbatim}

\begin{Verbatim}[breaklines,fontsize=\tiny,frame=single]
<classification1,DP word1,DP word2,DP word3>
\end{Verbatim}

\begin{Verbatim}[formatcom=\color{red},breaklines,fontsize=\tiny,frame=single]
Movie reviews are classified as either positive or negative. The movie review '{review}' is classified as '{classification1}'. Your task is to modify the review with minimal edits to flip the sentiment prediction. Please only produce the modified review and enclose it within <new> and </new> tags. Avoid making any unnecessary changes, and don't produce anything else.
\end{Verbatim}

\begin{Verbatim}[breaklines,fontsize=\tiny,frame=single]
<new>{counterfactual}</new>
\end{Verbatim}

\begin{Verbatim}[formatcom=\color{red},breaklines,fontsize=\tiny,frame=single]
Given the movie review '{counterfactual}', your task is to classify it either as 'positive' or 'negative' (and nothing else). Please only produce the classification and enclose it within <new> and </new> tags.
\end{Verbatim}

\begin{Verbatim}[breaklines,fontsize=\tiny,frame=single]
<new>{classification2}</new>
\end{Verbatim}

\begin{Verbatim}[formatcom=\color{red},breaklines,fontsize=\tiny,frame=single]
A counterfactual of a movie review x1, is another movie review x2 that is modified from x1 with minimal number of edits such that the classification of x2 is different from x1. Review your previous selection of the 3 most important words in x1 below with relation to its classification (also stated below), using the initial movie review x1 and the counterfactual movie review x2. If you are confident about your answer, maintain your answer. Otherwise, update your answer. Note that the words must only come from x1, and never from x2. Please follow the following output format and produce nothing else: [word1,word2,word3]
### initial movie review x1 with rating '{classification1}'
{review}
### counterfactual movie review x2 with rating '{classification2}'
{counterfactual}
### 3 most important words from x1
[DP word1,DP word2,DP word3]
\end{Verbatim}

\begin{Verbatim}[breaklines,fontsize=\tiny,frame=single]
[CF word1,CF word2,CF word3]
\end{Verbatim}

\paragraph{Prompts for obtaining decision-changing score:}

\begin{Verbatim}[formatcom=\color{red},breaklines,fontsize=\tiny,frame=single]
Given the movie review '{masked review}', for each sub-string '{MASK}' in the movie review, replace it with a word or short phrase such that the overall sentiment of the movie review becomes '{classification2}'. Also make sure that the sentence flows coherently. Please only produce the modified review and enclose it within <new> and </new> tags. Avoid making any unnecessary changes, never change any part of the review that is not '{MASK}', and don't produce anything else.
\end{Verbatim}

\begin{Verbatim}[breaklines,fontsize=\tiny,frame=single]
<new>{new review}</new>
\end{Verbatim}

\begin{Verbatim}[formatcom=\color{red},breaklines,fontsize=\tiny,frame=single]
Given the movie review '{new review}', your task is to classify it either as 'positive' or 'negative' (and nothing else). Please only produce the classification and enclose it within <new> and </new> tags (i.e., <new>positive</new> or <new>negative</new>).
\end{Verbatim}

\begin{Verbatim}[breaklines,fontsize=\tiny,frame=single]
<new>{new classification}</new>
\end{Verbatim}

\section{Prompt for IMDB Dataset} \label{app:IMDB_prompts}

We provide the prompts used for the IMDB dataset. The user's prompts are colored in red, and the LLM's outputs are colored in black. We present the case of $k=3$.

\paragraph{Prompts for DP:}

\begin{Verbatim}[formatcom=\color{red},breaklines,fontsize=\tiny,frame=single]
Given the movie review below, your task is to classify it either as 'positive' or 'negative', and state the top 3 most important words that led to you decision. Please follow the following output format and produce nothing else: <new>classification,word1,word2,word3</new>
### movie review
{review}
\end{Verbatim}

\begin{Verbatim}[breaklines,fontsize=\tiny,frame=single]
<new>classification1,DP word1,DP word2,DP word3,DP word4,DP word5</new>
\end{Verbatim}

\paragraph{Prompts for CFP:}

\begin{Verbatim}[formatcom=\color{red},breaklines,fontsize=\tiny,frame=single]
Given the movie review below, your task is to classify it either as 'positive' or 'negative' (and nothing else). Please only produce the classification and enclose it within <new> and </new> tags (i.e., <new>positive</new> or <new>negative</new>)).
### movie review
{review}
\end{Verbatim}

\begin{Verbatim}[breaklines,fontsize=\tiny,frame=single]
<new>{classification1}</new>
\end{Verbatim}

\begin{Verbatim}[formatcom=\color{red},breaklines,fontsize=\tiny,frame=single]
Movie reviews are classified as either positive or negative. The movie review below is classified as 'negative'. Your task is to modify the review with minimal edits to flip the sentiment prediction. Please only produce the modified movie review, and don't produce anything else.
### movie review
{review}
\end{Verbatim}

\begin{Verbatim}[breaklines,fontsize=\tiny,frame=single]
{counterfactual}
\end{Verbatim}

\begin{Verbatim}[formatcom=\color{red},breaklines,fontsize=\tiny,frame=single]
Given the movie review below, your task is to classify it either as 'positive' or 'negative' (and nothing else). Please only produce the classification and enclose it within <new> and </new> tags.
### movie review
{counterfactual}
\end{Verbatim}

\begin{Verbatim}[breaklines,fontsize=\tiny,frame=single]
<new>{classification2}</new>
\end{Verbatim}

\begin{Verbatim}[formatcom=\color{red},breaklines,fontsize=\tiny,frame=single]
A counterfactual of a movie review x1, is another movie review x2 that is modified from x1 with a minimal number of edits such that the classification of x2 is different from x1. Given a strings x1 and x2 below, your task is to study x1 and x2 carefully, and state the top 3 most important words (that are not stop words) from x1 that led to your previous decision of 'negative'. Please follow the following output format and produce nothing else: [word1,word2,word3]
### x1:
{review}
### x2:
{counterfactual}
\end{Verbatim}

\begin{Verbatim}[breaklines,fontsize=\tiny,frame=single]
[word1,word2,word3]
\end{Verbatim}

\paragraph{Prompts for CFS:}

\begin{Verbatim}[formatcom=\color{red},breaklines,fontsize=\tiny,frame=single]
Given the movie review below, your task is to classify it either as 'positive' or 'negative', and state the top 3 most important words that led to you decision. Please follow the following output format and produce nothing else: <new>classification,word1,word2,word3</new>
### movie review
{review}
\end{Verbatim}

\begin{Verbatim}[breaklines,fontsize=\tiny,frame=single]
<new>classification1,DP word1,DP word2,DP word3</new>
\end{Verbatim}

\begin{Verbatim}[formatcom=\color{red},breaklines,fontsize=\tiny,frame=single]
Movie reviews are classified as either positive or negative. The movie review below is classified as '{classification1}'. Your task is to modify the review with minimal edits to flip the sentiment prediction. Please only produce the modified movie review, and don't produce anything else.
### movie review
{review}
\end{Verbatim}

\begin{Verbatim}[breaklines,fontsize=\tiny,frame=single]
{counterfactual}
\end{Verbatim}

\begin{Verbatim}[formatcom=\color{red},breaklines,fontsize=\tiny,frame=single]
Given the movie review below, your task is to classify it either as 'positive' or 'negative' (and nothing else). Please only produce the classification and enclose it within <new> and </new> tags.
### movie review
{counterfactual}
\end{Verbatim}

\begin{Verbatim}[breaklines,fontsize=\tiny,frame=single]
<new>{classification2}</new>
\end{Verbatim}

\begin{Verbatim}[formatcom=\color{red},breaklines,fontsize=\tiny,frame=single]
A counterfactual of a movie review x1, is another movie review x2 that is modified from x1 with minimal number of edits such that the classification of x2 is different from x1. Review your previous selection of the 3 most important words in x1 below with relation to its classification (also stated below), using the initial movie review x1 and the counterfactual movie review x2. If you are confident about your answer, maintain your answer. Otherwise, update your answer. Note that the words must only come from x1, and never from x2. Please follow the following output format and produce nothing else: [word1,word2,word3]
### initial movie review x1 with rating '{classification1}'
{review}
### counterfactual movie review x2 with rating '{classification2}'
{counterfactual}
### 3 most important words from x1
[DP word1,DP word2,DP word3]
\end{Verbatim}

\begin{Verbatim}[breaklines,fontsize=\tiny,frame=single]
[CF word1,CF word2,CF word3]
\end{Verbatim}

\paragraph{Prompts for obtaining decision-changing score:}

\begin{Verbatim}[formatcom=\color{red},breaklines,fontsize=\tiny,frame=single]
Given the movie review below, for each sub-string '{MASK}' in the movie review, replace it with a word or short phrase such that the overall sentiment of the movie review becomes '{classification2}'. Ensure that your replacement makes the sentence flow coherently, but do not change anything else in the movie review (except for the '{MASK}' sub-strings). Please only produce the modified movie review, and don't produce anything else.
### movie review
{masked review}
\end{Verbatim}

\begin{Verbatim}[breaklines,fontsize=\tiny,frame=single]
{new review}
\end{Verbatim}

\begin{Verbatim}[formatcom=\color{red},breaklines,fontsize=\tiny,frame=single]
Given the movie review below, your task is to classify it either as 'positive' or 'negative' (and nothing else). Please only produce the classification and enclose it within <new> and </new> tags (i.e., <new>positive</new> or <new>negative</new>).
### movie review
{new review}
\end{Verbatim}

\begin{Verbatim}[breaklines,fontsize=\tiny,frame=single]
<new>{new classification}</new>
\end{Verbatim}

\end{document}